\documentclass[journal]{IEEEtran}
\ifCLASSINFOpdf
\else
\fi

\usepackage{epsfig}
\usepackage{graphicx}
\usepackage{amsmath}
\usepackage{amssymb}
\usepackage{bm}
\usepackage{multirow}
\usepackage{booktabs}
\usepackage{enumerate}
\usepackage{hyperref}
\usepackage{changes}
\usepackage{float}
\hypersetup{colorlinks}  
\usepackage[section]{placeins}
\usepackage{caption}
\usepackage{subcaption}
\usepackage{soul, xcolor}

\makeatletter
\newif\if@restonecol
\makeatother

\usepackage[linesnumbered,ruled,vlined]{algorithm2e}
\usepackage{algpseudocode}
\usepackage{amsmath}

\newcommand{\LL}[1]{\textcolor{black}{#1}}

\begin{document}
%

\title{Point Adversarial Self Mining: A Simple Method for Facial Expression Recognition}

\author{Ping Liu,
        Yuewei Lin,
        Zibo Meng,
        Lu Lu,
        Weihong Deng,
        Joey Tianyi Zhou*,
        and Yi Yang
\IEEEcompsocitemizethanks{\IEEEcompsocthanksitem P. Liu, Joey Zhou are with Institute of High Performance Computing, Agency for Science, Technology, and Research, Singapore. 
\IEEEcompsocthanksitem Y. Yang is with Centre for Artificial Intelligence, University of Technology Sydney, Sydney, Australia. 
\IEEEcompsocthanksitem W. Deng is with Pattern Recognition and Intelligent System Laboratory, Beijing University of Posts and Telecommunications, Beijing, China. 
\IEEEcompsocthanksitem Y. Lin is with Brookhaven National Laboratory, Upton, NY, USA.
\IEEEcompsocthanksitem Z. Meng is with InnoPeak Technology Inc., Palo Alto, CA, USA.
\IEEEcompsocthanksitem L. Lu is with Key Laboratory of Medical Molecular Virology, School of Basic Medical Sciences, Fudan University, Shanghai, China.
\IEEEcompsocthanksitem * Joey Tianyi Zhou is the corresponding author.
}

\thanks{Manuscript received Jan 22, 2021.}}

\markboth{Journal of \LaTeX\ Class Files,~Vol.~14, No.~8, August~2015}%
{Shell \MakeLowercase{\textit{et al.}}: Bare Demo of IEEEtran.cls for Computer Society Journals}

\IEEEtitleabstractindextext{%
\begin{abstract}

In this paper, we propose a simple yet effective approach, named Point Adversarial Self Mining (PASM), to improve the recognition accuracy in facial expression recognition. Unlike previous works focusing on designing specific architectures or loss functions to solve this problem, PASM boosts the network capability by simulating human learning processes: providing updated learning materials and guidance from more capable teachers. Specifically, to generate new learning materials, PASM leverages a point adversarial attack method and a trained teacher network to locate the most informative position related to the target task, generating harder learning samples to refine the network. The searched position is highly adaptive since it considers both the statistical information of each sample and the teacher network capability. Other than being provided new learning materials, the student network also receives guidance from the teacher network. After the student network finishes training, the student network changes its role and acts as a teacher, generating new learning materials and providing stronger guidance to train a better student network. The adaptive learning materials generation and teacher/student update can be conducted more than one time, improving the network capability iteratively. Extensive experimental results validate the efficacy of our method over the existing state of the arts for facial expression recognition.
\end{abstract}


}

\maketitle

\IEEEdisplaynontitleabstractindextext

\IEEEpeerreviewmaketitle


\section{Introduction}\label{sec:introduction}

\IEEEPARstart{F}{a}cial expression analysis aims to comprehend the underlying human emotions and establish efficient communications between humans and humans or humans and computers~\cite{zhong2014learning,wang2018thermal,zhao2015automatic,jang2018facial}. Due to its emerging applications in human-computer interaction, facial expression recognition (FER) has received massive interest among the research community. {In the past decade, the facial expression recognition accuracy has been boosted significantly with the rapid development of modern Convolutional Neural Networks (CNNs)~\cite{rodriguez2017deep}}.

As a ``data-hungry" method, CNNs {usually} require a huge amount of {annotated data for parameter learning}. Existing FER datasets can be categorized into two groups: lab-controlled and in-the-wild datasets. In lab-controlled datasets, such as CK+~\cite{lucey2010extended}, the collecting environment is highly controlled,~\textit{e.g.}, frontal exaggerated expressive faces with limited occlusions and minimal illumination changes. These lab-controlled datasets have been widely adopted for evaluating proposed methods~\cite{zhang1998comparison,zhang2005active,tian2002evaluation,eckhardt2009towards,yang2007boosting,hu2008multi,dahmane2011emotion,7025283,Liu2014,meng2017identity,xie2019adaptive}. The limited sample number and small variations in those lab-controlled datasets make it difficult to train a deep network with high generalities. To improve model generalization abilities, researchers collect data in a more challenging environment (in-the-wild), and annotate those in-the-wild data with facial expression labels. Compared to their lab-controlled counterparts, samples from in-the-wild datasets, which contain spontaneous head poses and various occlusions, can better reflect data distribution in the real world. Therefore, in-the-wild datasets for FER are attracting more research interests in the research community.

\begin{figure*}[htbp]
\begin{center}
\includegraphics[width=1.0\linewidth]{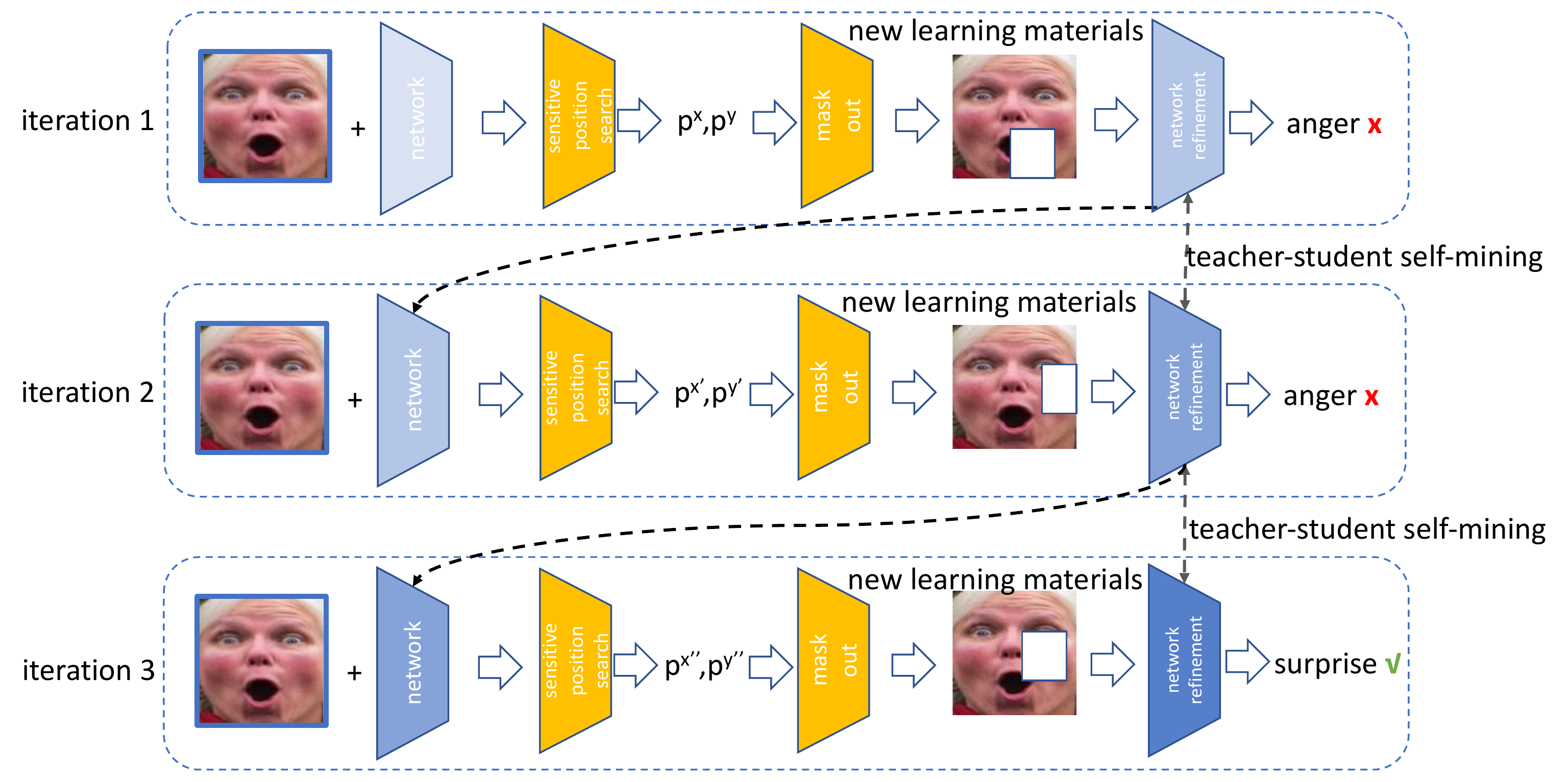}
\end{center}

\caption{{The pipeline of the proposed PASM. To improve network recognition rate on facial expression recognition, PASM keeps generating new learning materials for the network and providing guidance from an updated teacher network, which was a student in previous iterations. By taking advantage of new learning materials and updated teachers, PASM can progressively boost the recognition accuracy on facial expression datasets. Best viewed in color.}}
\label{fig:pipeline_SSPAM}
\end{figure*}

To improve performance of FER on those in-the-wild datasets, various CNNs based methods have been proposed~\cite{Zhang2020_ijcai,li2018occlusion,Liu2019d_arXiv}. Generally, those works manually design advanced architectures by utilizing an attention mechanism to simulate the human perception system. The attention map can be generated either by following a weakly supervised object localization strategy~\cite{Zhang2020_ijcai}, or by the occlusion score generated from an occlusion detector~\cite{li2018occlusion}. For example,~\cite{Zhang2020_ijcai} designs a weakly supervised local-global relation network to generate attention maps, indicating the facial regions crucial to the prediction. \cite{li2018occlusion} detects face landmarks and utilizes the landmark confidence scores as the corresponding occlusion level. Based on the confidence score, the network can explicitly place appropriate focus on different facial regions to make a more  accurate prediction. All those previous works need to introduce additional modules, increasing model sizes and inference complexity. 

From a different point of view, we consider to boost the facial expression recognition accuracy by simulating human learning process. Unlike previous works modifying model architectures, we improve the model capability by iteratively providing new learning materials and more advanced teachers. We argue that the same or even better performance can be achieved, if we appropriately utilize the training samples in hand. After all, the key information related to the target task already exists in the given samples.


To this end, we propose a self-mining framework to boost the recognition accuracy for FER and name it Point Adversarial Self Mining (\textbf{PASM}). The proposed PASM integrates the idea of point adversarial attack and teacher-student optimization strategy, making them collaborate together in an iterative way. In each iteration, the point adversarial attack module locates the most informative regions in each given sample. Because each sample has its own statistical distribution, the most sensitive region obtained by point adversarial attack is highly adaptive and has a direct influence on the classification. For each image, the located region is controlled by two factors: the statistical distribution of the image itself and the network for locating the region. By erasing the located region in the corresponding sample, a new image, which is treated as a ``harder"  learning material, is generated. Given those updated learning materials, the network has to spend more efforts to mine more knowledge to improve. In the following rounds/epochs, the improved network becomes an advanced teacher, providing guidance to train a better student network. Our PASM has three main steps: (1) train a teacher network on given samples; (2) locate the most informative region for each training sample by the teacher network trained in step (1), and then erase the located region to update learning {samples}; (3) train a student network on those new learning materials and under the guidance from the teacher network. The step (2) and step (3) can be conducted more than one time, by treating the student obtained in the previous iteration as a teacher in the current iteration. When the learning process finishes, the final student network is with high generalization ability, which is demonstrated in our experimental results. The whole iterative process is illustrated in Fig.~\ref{fig:pipeline_SSPAM}.

Our proposed PSAM is inspired by two previous related works: random erasing~\cite{zhong1708random} and adversarial erasing~\cite{wei2017object_cvpr}, where sub-regions were erased from the original ``clean" image to produce new samples to refine networks. However, there are a few significant differences between our method and ~\cite{zhong1708random, wei2017object_cvpr}: (1)~\cite{zhong1708random} randomly chooses the sub-regions to erase, without considering the characteristics of the input and the network prediction capability. In other words, it is possible that the selected region has little discriminative information, and erase them will not have any influence on the final prediction. On the contrary, our method behaves more reasonably since it locates the most informative region by considering the statistical information of each sample and the trained network capability; (2) unlike~\cite{zhong1708random} working as a pure data augmentation strategy, our method integrates harder sample generation and teacher-student optimization strategy into a unified self-mining framework. The two parts in our self-mining framework are conducted iteratively and work collaboratively; (3) compared to~\cite{wei2017object_cvpr} using an attention map to select regions, the region selected by our method is more structured and sparse, which aligns with the previous finding~\cite{Liu2014, zhong2012learning}: in face analysis problems, not all facial regions but only a few of them contribute to the final predictions.

\noindent

To sum up, our main contributions in this work are as follows:

(1) We propose a simple yet effective method, named Point Adversarial Self Mining (PASM), to improve the recognition accuracy of facial expression recognition. Compared to previous works~\cite{
ding2020occlusion,Wang_2020_CVPR,Wang2020} designing specific architectures, our method does not bring any additional parameters or computation cost in inference stages, while still achieves higher or comparable performance.

(2) Our method simulates the studying process in human societies. The method progressively generates new learning materials and provides guidance from a teacher network keeping updated. In PASM, the new learning materials are generated in an adaptive manner, considering both the statistical information of original learning materials and the teacher network status. Benefiting from those updated learning materials and teachers, the learning capability of the student network keeps improving.

(3) We conduct extensive experiments on challenging facial expression datasets collected under real-world settings. Our method, although simple, achieves better or comparable performance compared with the previous methods utilizing complex architectures or dedicated loss functions. 

\section{Related Work}

In this section, we will elaborate on previous works that are the most related to our work, including facial expression recognition approaches, data-augmentation strategies, and adversarial attack methods. 

\subsection{Facial Expression Recognition}
Facial expression recognition is an image-level classification research topic, which has been considered as a combination of three major steps: feature learning, feature selection, and classifier construction~\cite{Liu2014}. Before the dominance of convolutional neural networks in this field, hand-crafted feature-based methods have been exhaustively studied. As elaborated by the recent survey papers~\cite{zeng2008survey,sariyanidi2014automatic, zhang2017facial}, various hand-crafted features, such as Gabor-wavelet-based features~\cite{zhang2005active,tian2002evaluation,yang2007boosting},  Histograms of Oriented Gradients (HOGs)~\cite{hu2008multi,dahmane2011emotion}, Local Binary Pattern (LBP)~\cite{senechal2011combining,valstar2012meta,liu2014feature}, have been developed and well demonstrated for data collected under lab-controlled settings. 
In the past decade, as convolutional neural networks have shown promise in different computer vision tasks, researchers in facial expression recognition start to shift their attention from hand-crafted features to deep features~\cite{Liu2014,mollahosseini2016going,meng2017identity,Zhang_2018_CVPR,yang2018facial,li2018occlusion}. Comparing to their hand-crafted contemporaries, whose designing heavily depends on human expertise, deep features can be learned in a data-driven manner and have better performance on challenging datasets. To further improve the performance of CNNs 
for facial expression recognition, various architectures have been developed and studied, such as Deep Belief Networks~\cite{Liu2014}, ResNets~\cite{meng2017identity}, InceptionNets~\cite{mollahosseini2016going}, Generative Adversarial Networks (GANs)~\cite{Zhang_2018_CVPR,hu2019unsupervised,zhang2020face}. Most recently, researchers found that utilizing information from other modalities, such as brain waves~\cite{8283814}, audio~\cite{8392456,8695342}, can also boost emotion recognition performance. For a systematic review for deep learning in facial expression recognition, please refer to~\cite{Huang2020}.

\subsection{Data Augmentation}
Data augmentation is one of the strategies that can effectively prevent deep networks from over-fitting. As the architectures of the deep networks become deeper and more complicated, the number of parameters has increased dramatically. Without enough labeled data, it is easy for those networks to over-fit and lose the generality. In particular, as pointed out in~\cite{zhong1708random}, in an extreme case, an over-fitted model might achieve perfect accuracy for the training data while performs poorly on unseen data. To deal with the over-fitting problem, various data augmentation strategies have been proposed and employed~\cite{zhong1708random}. The basic idea of data augmentation is to introduce more variations into the training data without changing the statistical distribution of the original data. The most common and frequently used data augmentation techniques include random cropping, random flipping, and random color jittering. The efficacy of those three data augmentation strategies have been demonstrated in~\cite{krizhevsky2012imagenet,simonyan2014very}. In the past two years, two novel data augmentation methods, namely mixup~\cite{abs-1710-09412} and random erasing~\cite{zhong1708random}, have been proposed. Mixup~\cite{abs-1710-09412} combines pairs of examples and corresponding labels in a convex manner in order to generate new training samples. Random erasing~\cite{zhong1708random} randomly selects a position in a given input and erases pixels around the selected position. All of those proposed data augmentation methods are complementary and can be combined together to train a deep neural network, which has been experimentally proved~\cite{shorten2019survey}. 

There are significant differences between PASM and random erasing~\cite{zhong1708random}: 1) random erasing~\cite{zhong1708random} selects the position to erase in a purely random manner, while our method selects the erasing position by considering the data statistical information and network capabilities; {2)~\cite{zhong1708random} improves the model capability in a data augmentation manner, while our method works by simulating human learning process: providing new learning materials and advanced teachers, both of which are updated in an adaptive manner;} 3) our method works in a progressive way, continuously improving the model capability.

\subsection{Adversarial Attack}
With the successful application of deep neural networks in various computer vision tasks~\cite{9376704_heheefan_2021, luo2019taking, luo2018macro,luo2019significance,pan2020adversarial,9346018}, CNNs have been deployed in more safety-critical scenarios~\cite{luo2021category, luo2020ASM}, such as autonomous driving, financial fraud detection, etc. However, recent works~\cite{szegedy2013intriguing} pointed out that deep neural networks are not as stable as originally expected. On the contrary, they are vulnerable to adversarial examples, which are intentionally designed to mislead a trained deep neural network to make incorrect predictions. In previous works, adversarial examples are synthesized based on the model capability and given input~\cite{Rao2020}. 

Generally, adversarial attack can be categorized into two groups: white-box attack~\cite{goodfellow2014explaining,moosavi2016deepfool,kurakin2016adversarial} and black-box attack~\cite{Su2019,jiang2019black_mm}. The difference between them is the availability of network information,~\textit{e.g.}, parameters, gradients, etc. Specifically, in a white-box attack setting, those network information is available to the attackers; while in a black-box attack, it is unavailable to adversaries. Correspondingly, in a white-box attack, the model architecture and parameters are utilized to create the adversarial samples which can attack the model to the most, while in the black-box attack, a feed-and-query strategy is the first choice for adversaries. The most frequently used white-box attack policies include Fast Gradient Sign Method (FGSM)~\cite{goodfellow2014explaining}, Deep Fool~\cite{moosavi2016deepfool}, Projected Gradient Descent (PGD)~\cite{kurakin2016adversarial}, etc. Recently, researchers have conducted a few works toward black-box attack.~\cite{jiang2019black_mm} proposes to extend natural evolution strategies to estimate gradient for black-box image attack.~\cite{Papernot2017} studies the feasibility of utilizing the adversarial example transferability for black-box attack on static images. Readers can refer to~\cite{Rao2020} for a systematic review of the adversarial attack.

\section{Methodology}
\label{sec:method}
This section illustrates the details of the proposed PASM for facial expression recognition. As illustrated in Alg.~\ref{alg:SPPAM}, {at the first step, we train a network based on given training samples. The trained network is utilized to generate updated learning samples in the second step, acts as a teacher network in the third step. At the second step, the most sensitive position in each image is located by a point adversarial attack method, and then the local regions around the located position are masked out to generate an updated sample. At the third step, given the guidance from the teacher network, a student network is trained based on the updated learning materials generated in the second step. Step 2 and step 3 can be repeated more than one time, in which the student network obtained the previous epoch/iteration acts as an advanced teacher in the current epoch/iteration.}


\subsection{Initialize a Teacher Network}
\label{subsec:target_network_training}
{In the first step, we train a network based on the original training samples,~\textit{i.e.}, $\{x_i, y_i\}$, where $x_i$ denotes a sample and $y_i$ denotes the corresponding label. The target network is denoted as $F(w)$, where $w$ denotes the network parameters. In order to conduct a fair comparison, we choose ResNet-34~\cite{he2016deep} and VGG16~\cite{simonyan2014very} pretrained by ImageNet~\cite{ILSVRC15} as {an initial teacher} in our experiment. {In our training, we reset the output number to fit our interesting class number,~\textit{i.e.}, 7.} We choose a cross-entropy (CE) loss in this step, which is formulated as:}

\begin{equation}
    \min_{w}\frac{1}{N}\sum_{i}^{N} \mathrm{CE}((F(w, x_{i}),y_{i}))
\end{equation}
\label{eq:cross_entropy_target_network_training}
where $N$ denotes the number of training samples, $w$ denotes the network parameters.

\subsection{Sensitive Location Search via Point Adversarial Attack}
\label{subsec:point_adver_attack}
{Given the initial teacher network, we locate the most sensitive position in each training sample based on its statistical distribution and the network status. Two aspects need to be considered in this step. On the one hand, in facial activity analysis, previous studies~\cite{Liu2014, zhong2012learning} have shown that not all facial regions make equal contributions. As a matter of fact, only a sparse set of regions on faces contain important information and contribute to the final prediction~\cite{zhong2012learning,Liu2014,liu2014feature}. On the other hand, different images have different statistical distributions, and therefore the key region(s) in each image probably differ. In summary, the location search should be \textit{selective} and \textit{adaptive}.}

{To this end, we utilize point adversarial attack in this step. As an adversarial attack method, point adversarial attack can locate an image region that is sensitive to final predictions. The location searching process is adaptive since it is dependent on two factors: the statistical distribution of the image itself, and prediction capability of the network used to search the location. Unlike previous unrestricted adversarial attack works such as~\cite{goodfellow2014explaining}, point adversarial attack searches one solo position for each given image, which not only complies with the findings in~\cite{Liu2014, zhong2012learning,JMLR:v20:13-580} but also benefits to the interpretability.}

The general goal of adversary attack can be formulated as follows:
\begin{equation}
\begin{aligned}
    && \max_{e({\bold{x}})^{*}} F_{adv}(\bold{x} + e(\bold{x}), w) \\
    && s.t. \qquad c(e(\bold{x}))
    \end{aligned}
    \label{eq:point_adver_attack}
\end{equation}
where $F(*, w)$ denotes the target network, $\bold{x}$ denotes the original input without perturbations, $e(\bold{x})$ is an ``additive adversarial perturbation"~\cite{Su2019} with respect to $\bold{x}$, $F(\bold{x},w)$ is the prediction for $\bold{x}$, $F(\bold{x} + e(\bold{x}), w)$ is the prediction for the perturbed input,~\textit{i.e.}, $\bold{x}+e(\bold{x})$. {Adversarial attack aims to find perturbations under a specified constraint,~\textit{i.e.}, $c(e(\bold{x}))$, to mislead the network $F(*,w)$. In our case, the constraint $c(e(\bold{x}))$ is $||e(\bold{x})||_{0} \leq d$, in which $d$ is set as one to make the perturbation applied on one solo position.}

{We follow the point attack method proposed in~\cite{Su2019} to locate the informative position in an adaptive manner.} The key part of the point attack calculation is based on differential evolution, rather than gradient descent/ascent. {Starting from a population of solution candidates, each of which denotes a perturbation and is encoded as a vector, a differential evolution method conducts population selection and inheritance to generate better solutions for the target. Concretely, each solution candidate,~\textit{i.e.}, $\hat{x}_{*}()$, encodes the spatial coordinates to apply the point attack. At each iteration(generation), a new candidate solution is produced by the following formulation:}

\begin{equation}
    \hat{x}_{i}(g+1)=\hat{x}_{r1}(g) +k(\hat{x}_{r2}(g) - \hat{x}_{r3}(g))
   \label{eq:point_adver_attack_2}
\end{equation}
{where $g$ is the generation index, $\hat{x}_{i}$ is a candidate solution, $r1$, $r2$, $r3$ is candidate index in the same generation~\footnote{$r1\neq r2 \neq r3$}, $k$ is a predefined scale factor. A new candidate solution in generation $g+1$ is generated by three different solution candidates selected in generation $g$. The candidate solution will compete with their parents and it will be saved for further calculation only if it is better than its parents.}

For each training sample, we use the point adversarial attack method discussed above to calculate an attack sample $\hat{x}_{*}()$, which contains the attack position denoted as ($p^{x}, p^{y}$). The calculation process needs to consider the statistical information of the input and the prediction capability of network $F(*, w)$. {For different samples, since they have different statistical information, the calculated attack positions for each image are different; for each sample, if we search the attack position using different networks,~\textit{e.g.}, $F(*, w)$ saved in different epochs, the located attack positions are also different. Therefore, the sensitive positions located in our method meet the original requirements: selective and adaptive.}

\subsection{{Generating New Learning Materials to Learn}}
\label{subsec:augmented_sample_generation}
In this step, we utilize the sensitive positions located in the previous step,~\textit{i.e.}, ($p^{x}, p^{y}$) to generate {learning materials} for network refinement. {The new learning materials are generated} by erasing the local information centered at the located position ($p^{x}, p^{y}$), since the perturbations added at those located positions are easy to mislead the network. {This step is like providing new textbooks for students in each semester, based on the learning capability of students.}

Concretely, for each given sample $\bold{x}_{i}$, assume its size is $W \times H$, where $W$ denotes the width and $H$ denotes the height, the sensitive position located in Section.~\ref{subsec:point_adver_attack} is ($p_{}^{x}, p_{}^{y}$), we mask out the information in a local region centered at ($p_{}^{x}, p_{}^{y}$) with a size of $s \times s$. The generated samples are used as new learning materials to improve the capability of network.

\begin{algorithm}
  \caption{Point Adversarial Self Mining}
  \label{alg:SPPAM}
  \KwIn{input image set $\{\bold{x}_{i}, y_{i}\}$, $1 \leq i \leq N$; image sizes $W_{i}$ and $H_{i}$; patch size $S$; iteration number $iter$}

  \KwOut{betwork parameter $w^{*}$
  }
  {\textbf{initialize a teacher network:}} \:
  
  training target network with input images $\{\bold{x}_{i}, y_{i}\}$, get an initialization $w$ for a teacher network \;
  \For{$ind = 1, ..., iter$}
  {
    \For{i = 1, ..., N}
    { {\textbf{generating new learning materials:}} \:
    
      sample an image $\{\bold{x}_{i}, y_{i}\}$ from the given set\;
      for the sampled $\{\bold{x}_{i}, y_{i}\}$, utilize Equation.~\ref{eq:point_adver_attack_2} to generate the sensitive position in it\;
      mask out the region centered at the searched location \;
          set the pixel value in the masked out region by the perturbed RGB value, generating a new learning sample $\bold{x}_{i}^{new}$\;
    
    {\textbf{refine a student network and update the teacher network:}} \:
    
    leverage the new learning sample $\{\bold{x}_{i}^{new}, y_{i}\}$ and the teacher network, refine a student network  based on Equation.~\ref{eq:teacher_student};\
    }
    set the student network as a teacher network for the next iteration.\
  }
  return $w_{ind}$ as $w^*$\;
\end{algorithm}

\subsection{{Update New Teacher to Guide Student}}
 
To train a network with better generality, we not only utilize the new learning materials, denoted as $\bold{x}'$, and also a teacher network to provide guidance during the learning.

\begin{figure}[htbp]
  \centering
  \includegraphics[width=1.0\linewidth]{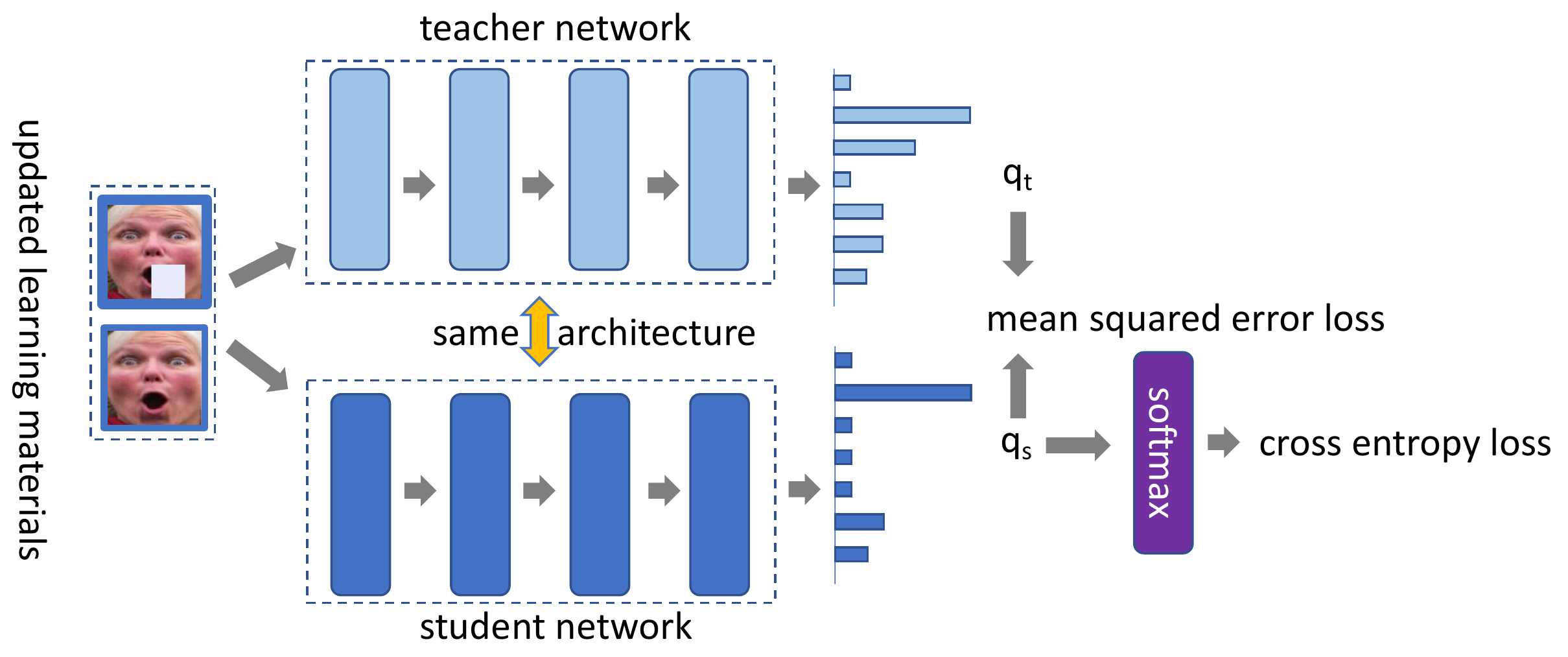}  
  \caption{ {The network refinement by updated learning samples and updated teachers. Unlike previous teacher-student works, in which the student network is with a smaller size, our teacher and student network share the same architecture. The deeper color, the better performance. Best viewed in color. }}
  \label{fig:teacher-student}
\end{figure}

Specifically, as illustrated in Fig.~\ref{fig:teacher-student}, {the network used to locate the attack information acts as a teacher network now, providing guidance to learn a new network,~\textit{i.e.}, a student network with the same architecture.} We denote the student network as $F_{s}(*, w_s)$. The teacher network is fixed in the current round and provides guidance to train the student network. For each sample $\bold{x}_i$, the teacher network provides logits signal denoted as $q_{i}^{t}$. The logits $q_{i}^{t}$ generated by the teacher network, as well as the corresponding one-hot vector label $y_{i}$, are both utilized to supervise the student network training. Therefore, other than new learning samples, the student network receives additional guidance from the teacher network. To supervise the student network learning, we use a cross-entropy loss to minimize the discrepancy between the prediction and the ground truth hard label, and a mean squared error (MSE) loss to minimize the discrepancy between the logits from the teacher network and the student network. The formulation is as follows:

\begin{equation}
\begin{aligned}
L = \alpha * CE(F_{s}(x_{i}', w_s),y_{i}) \\
   + \beta * MSE(q_{i}^{s}(w_s, x_{i}'), q_{i}^{t}(w_t,x_{i}'))
\end{aligned}
\label{eq:teacher_student}
\end{equation}
where $F_s(*, w_{s})$ denotes the student network with parameter $w_{s}$, $q_{i}^{s}(x_{i}', w_s)$ denotes the logits generated by the student network,  while $q_{i}^{t}(x_{i}', w_t)$ denotes the logits generated by the teacher network, $\alpha$ and $\beta$ are parameters to balance the contribution of the two loss terms.

There are two advantages in our teacher-student training strategy: 1) the updated learning samples and the teacher-student optimization assist the network to better model inter-class variations in facial expression recognition~\cite{Zheng_2019_CVPR}; 2) unlike previous teacher-student learning works such as~\cite{44873}, our teacher network and student network share the same structure, relieving us from designing or selecting a proper teacher network and student network. We name our learning strategy as self-mining mechanism.


\begin{figure*}[htbp]
\begin{center}
\includegraphics[width=1.0\linewidth]{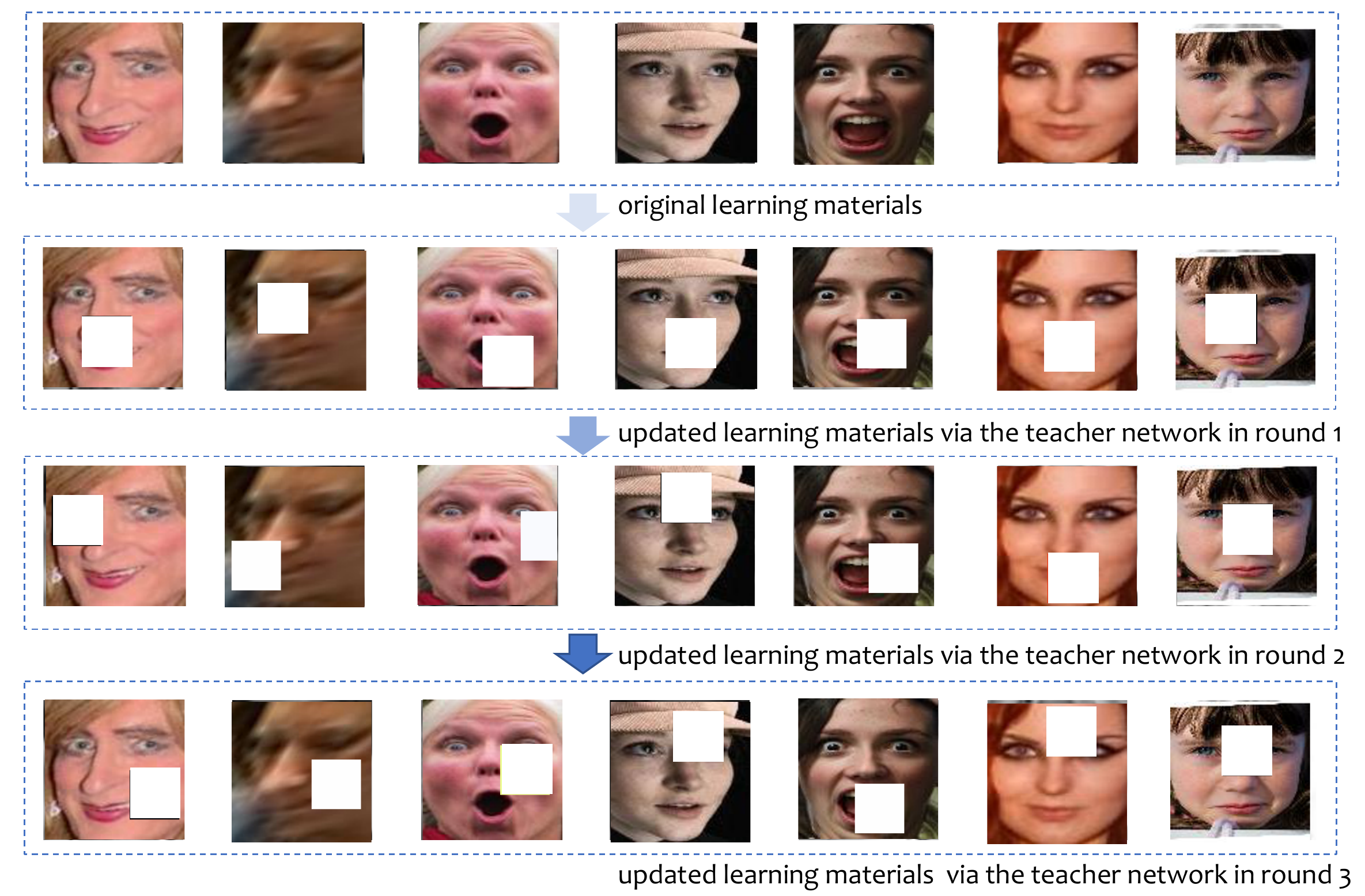}
\end{center}

\caption{{Visualization of updated learning images by PASM. Images in each row are new learning materials generated by an updated teacher network, which was a student in the previous iteration(row). Best viewed in color.}}
\label{fig:augmented_results_SSPAM_RAF}
\end{figure*}

\textbf{Iterative Mining Mechanism.} The aforementioned step can be conducted in an iterative way. In each iteration, the student network trained in the previous iteration changes its role, and becomes a teacher network in the current iteration. {Since the network capability keeps improving as the iteration increases, the located sensitive position for each image is different and the generated new learning samples vary. The process is illustrated in Fig.~\ref{fig:augmented_results_SSPAM_RAF}. With the help of updated new learning materials and a stronger teacher network, the student network capability keeps improving correspondingly. The details of our algorithm can be found in Alg.~\ref{alg:SPPAM}.}

\section{Experimental Results And Discussions}
\label{section:experiments_results}

{In this section, we utilize in-the-wild datasets,~\textit{i.e.}, Real-world Affective Face Database (RAF-DB) 2.0~\cite{li2018reliable}, FER2013~\cite{goodfellow2015challenges}, Occlusion-RAF-DB\cite{Wang2020}, Pose-RAF-DB\cite{Wang2020}, and one lab-controlled dataset,~\textit{i.e.}, Extended CohnKanade (CK+)~\cite{lucey2010extended} to demonstrate the efficacy of our method. All the details about these datasets and evaluation settings are described in the following subsections.}

\subsection{Experimental Setup}
We test our method on three frequently used facial expression datasets,~\textit{i.e.}, FER2013~\cite{goodfellow2015challenges}, Real-world Affective Face Database (RAF-DB) 2.0~\cite{li2018reliable}, Extended CohnKanade (CK+)~\cite{lucey2010extended}. RAF-DB and FER2013 are collected in the wild, covering variations in the real world, such as lighting conditions, large head pose, etc. To demonstrate the efficacy of our method when facing occlusion and pose issues in real scenarios, we further conduct experiments on Occlusion-RAF-DB~\cite{Wang2020} and Pose-RAF-DB~\cite{Wang2020}, which are constructed subsets from RAF-DB with additional occlusion and pose annotations.

    \textbf{RAF-DB}~\cite{li2018reliable} is a dataset collected from the internet. There are 29,672 facial images in this dataset. RAF-DB consists of highly diverse samples and covers different variations in the real world. The labels in this dataset are manually achieved by crowd-sourced annotation~\cite{li2018reliable}. This dataset has two kinds of annotations: basic expressions and compound expressions. To make comparisons with previous works, we only utilize the basic expression label set to test our method. For the basic expression label set, there are 12,271 images for training and 3,068 images for testing.
    
     \textbf{FER2013}~\cite{goodfellow2015challenges} is another widely used in-the-wild dataset. There are 28,709 samples in the training set and 3,589 samples in the testing set. All those images are collected by Google search engine and labelled with basic expressions. It was constructed for the ICML 2013 Challenges in Representation Learning. It should be noted that the original image size in this dataset is only $48 \times 48$. The small spatial size and high variations make the analysis difficult.

      \textbf{CK+}~\cite{lucey2010extended} is a dataset collected in a lab controlled environment. The original data collected is video-based, which has 593 video sequences in total. In the first frames of each sequence, the subject activates a neutral expression and gradually shifts to a peak expression in the last frames. There are 327 sequences with basic expression labels in this dataset. Unlike RAF-DB and FER2013, there is no official training/validation/test split. Previous works usually utilize the first frame as a neutral sample and the last three frames with the target expression labels to construct the data. In the constructed data, there are 1,308 images labelled with seven basic expressions in total.

      \textbf{Occlusion-RAF-DB and Pose-RAF-DB}\cite{Wang2020} are occlusion test subsets extracted from RAF-DB. In those subsets, there are various occlusion types, such as masks and glasses on faces, objects before faces, etc. Those constructed subsets are effective to test the efficacy of FER networks when facing real scenarios. Existing FER methods have a drop in recognition accuracy when testing in real cases full of occlusion and pose-variations~\cite{Wang2020}.
      
\begin{figure*}[htb]
\begin{center}
\includegraphics[width=1.0\linewidth]{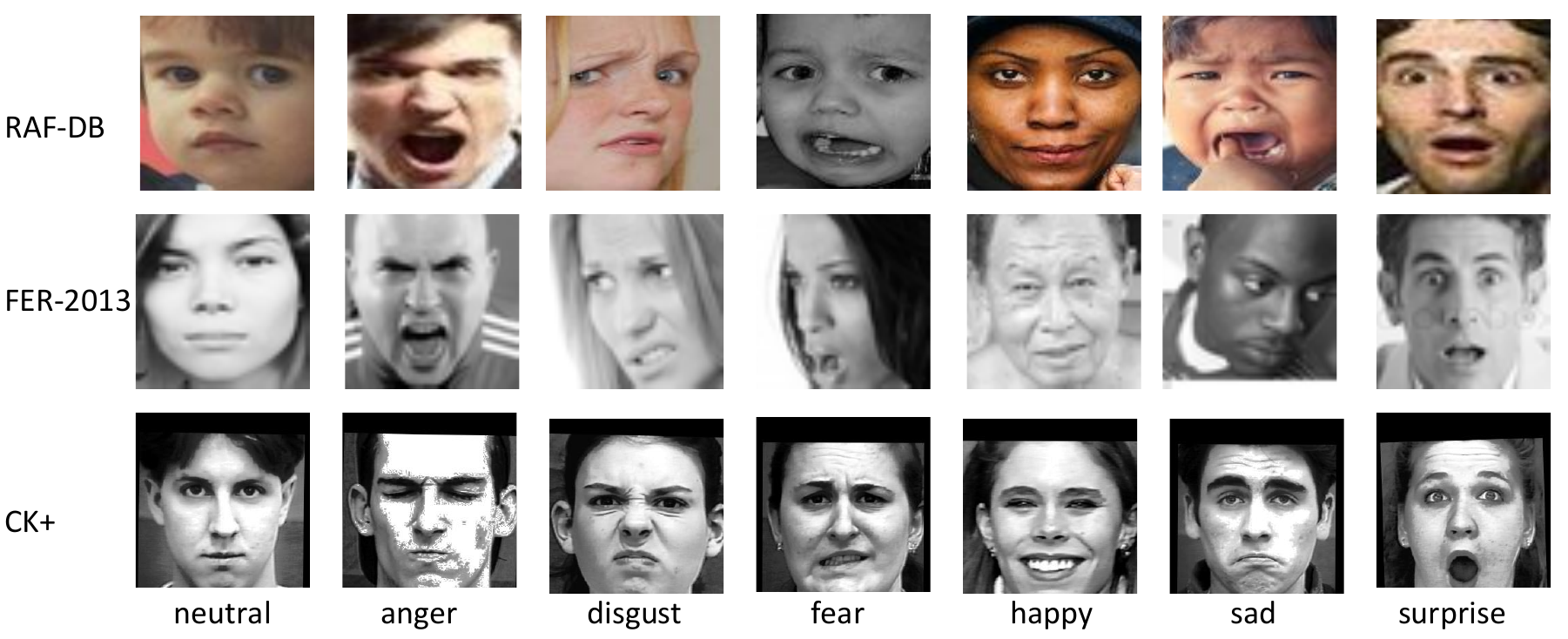}
\end{center}

\caption{Sample visualization for datasets utilized in this work,~\textit{i.e.}, RAF-DB, FER2013, CK+. From left to right: neutral, anger, disgust, fear, happy, sad, and surprise. From top to bottom: RAF-DB 2.0, FER-2013, and CK+. RAF-DB 2.0 and FER-2013 are in-the-wild databases, while CK+ is a dataset collected in controlled environments. Best viewed in color.}
\label{fig:visualization-databases}
\end{figure*}

Examples from those three datasets are shown in Fig.~\ref{fig:visualization-databases}. There are different variations in those datasets, including: 1) scale: the spatial size of each sample in FER2013 is $48 \times 48$, which are much smaller than that in RAF-DB and CK+; 2) lighting: the lighting conditions when capturing those datasets are quite different; 3) pose: there are heavy pose variations in RAF-DB and FER-2013, which are closer to the real scenarios.

To demonstrate the efficacy of our method, we utilize those aforementioned datasets for three evaluation settings,~\textit{i.e.}, inner-database evaluation, cross-database evaluation, occlusion-subset evaluation. The details of the three settings are illustrated as follows:

    \textbf{Inner-database Evaluation} In inner-database evaluation, we train the network on the training set of a dataset and test the trained network on the testing set of the same dataset. We conduct an inner-database evaluation on RAF-DB and FER2013.
    
    \textbf{Cross-database Evaluation} In cross-database evaluation, we train the network on the training set of a dataset and test the trained network on a different dataset. To make a fair comparison with previous work~\cite{Li2020}, in this setting, we conduct the training on RAF-DB and test the trained network on FER2013 and CK+.
    
    \textbf{Occlusion and Pose Subset Evaluation} In this setting, we train the network on RAF-DB by our method, and test it on the Occlusion-RAF-DB and Pose-RAF-DB, demonstrating the efficacy of our proposed method when dealing the challenging cases.

\subsection{Implementation Details}
To preprocess the data, we detect faces and conduct the alignment based on MTCNN~\cite{7553523} from each given image. All the aligned detected faces are resized to $224 \times 224$. For making fair comparisons with previous works, we choose ResNet-34 and VGG-16 as our backbones, which are pre-trained on ImageNet Dataset~\cite{deng2009imagenet}. The output number for both networks are reset to seven, corresponding to the number of basic expressions. We optimize our networks by stochastic gradient descent with 0.9 momentum. We set the initial learning rate to 0.01, which will be multiplied by 0.1 every ten epochs.  We implement all the experiments in PyTorch~\cite{NEURIPS2019_9015} and run our experiment on  NVIDIA GTX 2080Ti GPU cards.

\subsection{Performance Evaluation}
In this section, we test the discriminative ability of our method by conducting experiments for inner-dataset evaluations, and test the generality capability of our method by conducting experiments for cross-dataset evaluations. Our experiment is evaluated via mean classification accuracy.

\subsubsection{Inner-dataset Evaluations}
\label{sec:inner-dataset}

In this section, we utilize two in-the-wild databases,~\textit{i.e.}, RAF-DB, FER2013, to demonstrate the efficacy of our method. The result comparisons on the two datasets are reported in Table.\ref{table:inner-RAF} and \ref{table:inner-FER-2013}, respectively.
\textbf{Result Comparison on RAF-DB} We compare our method to previous state-of-the-art methods on RAF-DB and report the comparison in Table.~\ref{table:inner-RAF}. In Table.~\ref{table:inner-RAF},~\cite{cai2018probabilistic} introduces more manual labels into training and therefore needs more label cost,~\cite{ding2020occlusion} and~\cite{Wang2020} design region attention branch network, placing different weights on facial regions based on their occlusion conditions. Comparing to those previous works~\cite{cai2018probabilistic,ding2020occlusion,Wang2020}, our method achieves higher performance on both chosen architectures,~\textit{i.e.}, ResNet-34 and VGG-16. {To fairly compare with the latest works~\cite{Wang_2020_CVPR, Wang2020}, we test our method on ResNet-18 and still outperforms~\cite{Wang_2020_CVPR, Wang2020}.  Unlike those previous works, our method does not modify architectures or introduce any external data, bringing few additional computational or memory cost in the inference stage. }

\begin{table}[htbp]
\caption{Inner-dataset comparison on RAF-DB. }

\label{table:inner-RAF}
\centering
\begin{tabular}{lcccc}
\toprule
Method                    &Year & {Backbone} & Accuracy       \\ \midrule
FSN\cite{zhao2018feature}      & 2018  & {AlexNet} & $81.10\%$              \\
MRE-CNN\cite{fan2018multi}      & 2018  &{VGG-16} & $82.63\%$             \\

PAT-VGG-F-(gender,race)\cite{cai2018probabilistic}   & 2018 &{VGG-16}  &$83.83\%$ \\
PAT-ResNet-(gender,race)\cite{cai2018probabilistic}    & 2018 & {ResNet-34} &$84.19\%$ \\
OADN\cite{ding2020occlusion}  & 2020  &{ResNet-50} & $87.16\%$ \\
SCN\cite{Wang_2020_CVPR}  & 2020 &{ResNet-18} & $87.03\%$\\
RAN\cite{Wang2020} &2020 &{ResNet-18} &$86.90\%$\\
 \midrule
  PASM & 2020 &{VGG16} & $\textbf{87.50}\%$\\
  PASM & 2020 &{ResNet-18} & {$\textbf{87.18}\%$}\\
PASM (3 rounds) & 2020 &{ResNet-34} & $\textbf{88.68}\%$\\
\bottomrule
\end{tabular}
\end{table}

\textbf{Result Comparison on FER2013} We compare our method to previous state-of-the-art methods on FER2013 and report the comparison in Table.~\ref{table:inner-FER-2013}. FER2013 dataset was constructed for a facial expression recognition challenge. The small size of each sample and high variations introduced by pose and light conditions make the prediction rate much lower than RAF-DB. To improve the prediction accuracy on this dataset, various methods have been proposed. For example,~\cite{cai2018probabilistic} introduces an attribute tree convolutional neural network to explicitly model the facial attributes, such as race, gender, and age. From Table.~\ref{table:inner-FER-2013}, we can find that our method achieves a higher or comparable accuracy on this challenging dataset.

\begin{table}[htbp]\caption{Inner-dataset comparison on FER-2013.}

\label{table:inner-FER-2013}
\centering
\begin{tabular}{lccc}
\toprule
Method                  &Year &{Backbone} & Accuracy       \\ \midrule 
Guo et al.\cite{guo2016deep} & 2016 &{InceptionNet} & $71.33\%$\\
ECNN\cite{wen2017ensemble}    &2017  &{Ensemble}   & $69.96\%$             \\
Ron et al.\cite{breuer2017deep}      &2017  &{3 Conv} & $72.1\%$               \\

PAT-VGG-F-(gender,race).\cite{cai2018probabilistic} &2018 &{VGG-16}  &$72.16\%$ \\
PAT-ResNet-(gender,race).\cite{cai2018probabilistic}  &2018 &{ResNet-34} &$72.00\%$ \\
 \midrule
 PASM &2020  &{VGG-16} & $\textbf{72.73}\%$\\
 PASM (2 rounds) &2020 &{ResNet-34}  & $\textbf{73.59}\%$\\
 





\bottomrule
\end{tabular}
\end{table}

\subsubsection{Evaluations on Occlusion-RAF-DB and Pose-RAF-DB}
Occlusion-RAF-DB and Pose-RAF-DB are two test subsets with occlusion and pose annotations. Those two datasets are constructed to test the network capability under occlusion and pose variations.~\cite{Wang2020} designs a region attention network (RAN) which adaptively assigns different weights to each facial region based on their contributions. Other than that, a region biased loss is proposed.{ As shown in Table.~\ref{table:occlusion-pos-evaluation}, our method,~\textit{i.e.}, PASM, outperforms~\cite{Wang2020} on all three subsets. Specifically, for the occlusion subset, the gain is $0.55\%$; for pose larger than 30 degrees and 45 degrees, the gains are $2.92\%$ and $2.97\%$, respectively. Compared to Random Erasing~\cite{zhong1708random}, which does not have an \textbf{iterative} self-mining mechanism, PASM achieves higher performance on all three subsets. The better performance of our method on those three challenging datasets demonstrates that the better performance of our method comes from continuously updated new learning materials and advanced teacher networks.}

\begin{table}[htbp]\setlength{\tabcolsep}{13pt}
\caption{Comparison on Occlusion-RAF-DB and Pose-RAF-DB.}
\label{table:occlusion-pos-evaluation}
\centering
\begin{tabular}{lccc}
\toprule
RAF-DB      & Occlusion           &Pose(30)      & Pose(45)       \\ \midrule 
RAN~\cite{Wang2020} &$82.72\%$    &$86.74\%$  & $85.20\%$             \\ 
{RanEra~\cite{zhong1708random}}    & {$78.78\%$}    & {$85.00\%$}  & {$83.69\%$}             \\ \midrule  
Our method     &{$\textbf{83.27}\%$}   & {$\textbf{89.66}\%$}  &{$\textbf{88.17}\%$} \\
\bottomrule
\end{tabular}
\end{table}

\subsubsection{Cross-dataset Evaluations}
\label{sec:cross-dataset}
To test the generality of our method, we conduct a cross-dataset evaluation. We train the network on RAF-DB and test it directly on CK+ and FER2013. The comparisons with previous works are reported in Table.~\ref{table:cross-dataset-ck}-\ref{table:cross-dataset-fer2013}. FER2013, as discussed in previous sections, is a challenging dataset collected in the real world; CK+ is collected in a lab-controlled situation and is the most representative in-the-lab dataset. We want to test the generality of our method 
on those two datasets. {Again, by taking advantages of new learning materials and strong guidance provided from updated teacher networks, our method achieves a higher cross-dataset accuracy on CK+, a comparable accuracy on FER2013.}

\begin{table}[htp]\setlength{\tabcolsep}{9pt}

\caption{Cross-dataset comparison on CK+. }

\label{table:cross-dataset-ck}
\centering
\begin{tabular}{lllc}
\toprule
Method      & Source           &Target      & Accuracy       \\ \midrule 

CNN-Li \cite{Li2020}    &RAF-DB    &CK+  & $78.00\%$             \\ \midrule
Our method     &RAF-DB  &CK+   &$\textbf{79.65}\%$\\


\bottomrule
\end{tabular}

\end{table}


\begin{table}[htbp]\setlength{\tabcolsep}{11pt}

\caption{Cross-dataset comparison on FER2013. }


\label{table:cross-dataset-fer2013}
\centering
\begin{tabular}{lllc}
\toprule
Method      & Source           &Target      & Accuracy       \\ \midrule 

CNN-Li \cite{Li2020}    &RAF-DB    &FER2013  & $55.38\%$             \\ \midrule
Our method      &RAF-DB   &FER2013  & $\textbf{54.78}\%$ \\


\bottomrule
\end{tabular}

\end{table}


\subsection{Ablation Study}

We conduct extensive ablation studies in this section. At first, we analyse the generality of the self-mined knowledge; at second, we conduct a comparison with the random erasing and adversarial erasing for demonstrating the superiority of our self-mined strategy; at third, we analyse the relationship between the self-mining iteration number and the final accuracy; at fourth, we illustrate the impact of the size of erasing mask and parameter in Equation.~\ref{eq:teacher_student} to the final performance; last but not least, we conduct an evaluation of using random values for erasing.

\textbf{Generality of the Mined Knowledge}
{We analyze the generality of self mined knowledge extracted by our method and report the results in Table.~\ref{table:study-of-generality}. ``Mining Network" means the network used to locate a sensitive position to generate new learning samples, ``Training Network" denotes the network for extracting knowledge from learning materials. As shown in Table.~\ref{table:study-of-generality}, when ``Training Network" is different from ``Mining Network", the prediction rate drops a bit on both datasets. On RAF-DB, the performance drops from $87.54\%$ to $87.09\%$; on FER2013, the prediction rate drops from $73.50\%$ to $72.73\%$. We believe this performance drop comes from the difference between the mining network and the training network,~\textit{e.g.}, network structures, parameters. However, even when we use two different networks for self-mining and training, our performance on RAF-DB and FER2013 still outperforms or is comparable with previous works,~\textit{i.e.},~\cite{cai2018probabilistic} and ~\cite{ding2020occlusion}.}

\textbf{Comparison with Random Erasing and Adversarial Erasing}
We make a comparison with the random erasing~\cite{zhong1708random} and adversarial erasing~\cite{wei2017object_cvpr}. Since random erasing~\cite{zhong1708random} does not have an iterative self-mining mechanism,{~\textit{e.g., without updating learning materials or updating teacher networks  working collaboratively}}, we do not utilize teacher-student optimization and iterative refinement in this experiment (PASM w/o teacher). The comparison results are reported in Table.~\ref{table:random-erasing}. We can find that our method outperforms random erasing and adversarial erasing. {Compared to random erasing conducting erasing in a random position, our method locates the position in a more adaptive manner by both considering the statistical information of each sample and the network capability. The lower performance of Random Erasing~\cite{zhong1708random} compared to Adversarial Erasing~\cite{wei2017object_cvpr} also demonstrates that an adaptive manner outperforms a random way. Compared to adversarial erasing, our method achieves better accuracy. We erase the sub-region located in a more structured and sparse way,~\textit{i.e.}, a solo position, which aligns with the findings in previous works. }

\begin{table}[htbp]

\caption{Comparisons with Random Erasing and Adversarial Erasing. ResNet-34 is chosen as the backbone and trained for only one round. }

\label{table:random-erasing}
\centering
\begin{tabular}{lccc}
\toprule
Dataset          &Random Erasing   &Adv Erasing    & PASM w/o teacher      \\ \midrule 

RAF-DB       & $85.93\%$ &$86.57\%$  & $\textbf{86.86}\%$            \\ 
FER2013       & $72.92\%$ &$73.08\%$ & $\textbf{73.24}\%$             \\  


\bottomrule
\end{tabular}

\end{table}

\begin{figure}[htbp]
\begin{subfigure}{.5\textwidth}
  \centering
  \includegraphics[width=1.0\linewidth]{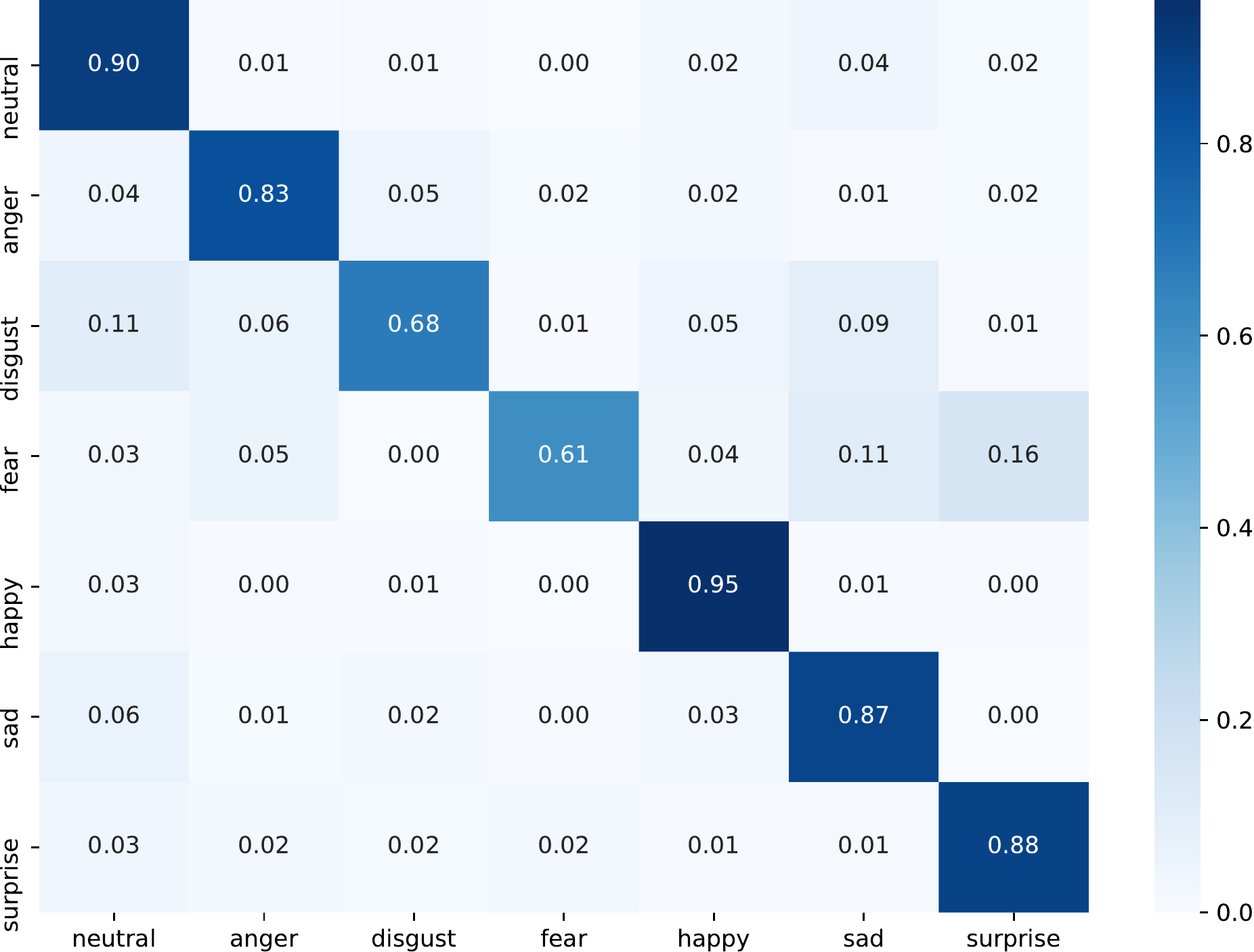}  
  \caption{Confusion Matrix of PASM on RAF-DB. }
  \label{fig:sub-first}
\end{subfigure} \qquad


\begin{subfigure}{.5\textwidth}
  \centering
  \includegraphics[width=1.0\linewidth]{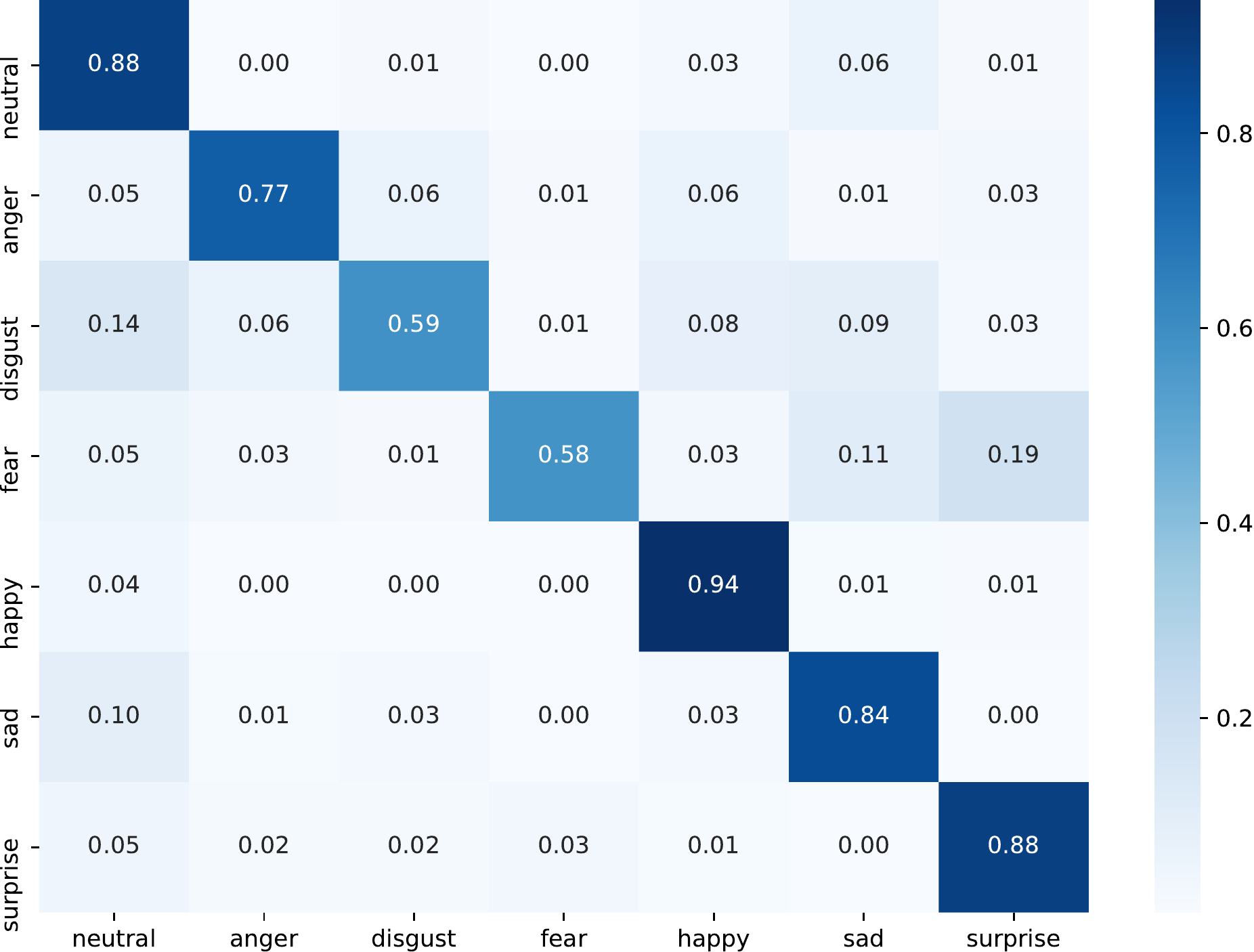}  
  \caption{Confusion Matrix of Random Erasing on RAF-DB. }
  \label{fig:sub-third}
\end{subfigure}

\caption{Confusion Matrix of PASM,  Random Erasing on RAF-DB. The darker the color, the higher the accuracy.}
\label{fig:confusion_matrix}
\end{figure}

We show the confusion matrix for PASM and Random Erasing in Fig.~\ref{fig:confusion_matrix}. It can be observed that PASM outperforms Random Erasing especially on ``anger" and ``disgust".

\begin{figure}[t]
\begin{subfigure}{.5\textwidth}
  \centering
  \includegraphics[width=1.0\linewidth]{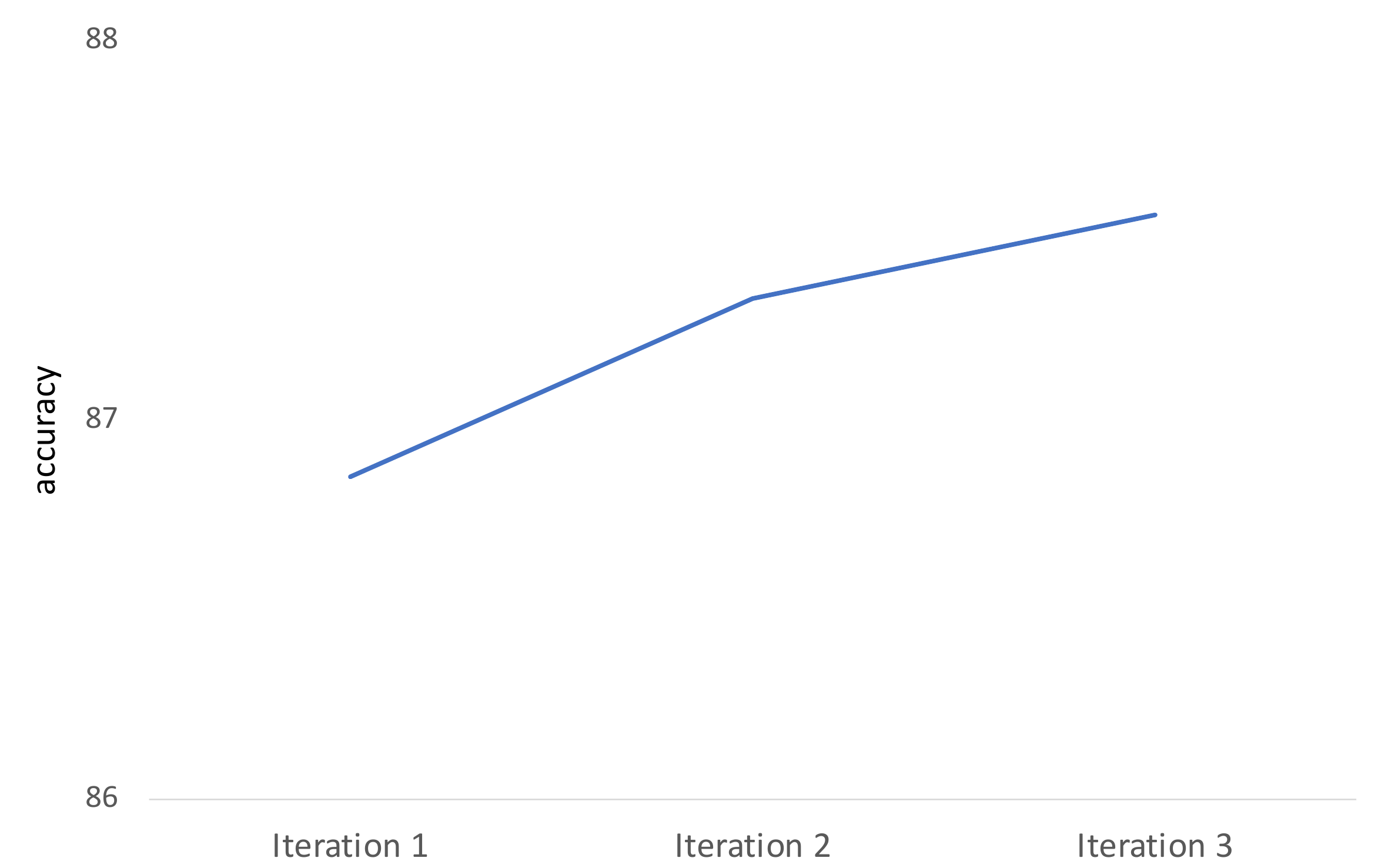}  
  \caption{ResNet-34 on RAF-DB. }
  \label{fig:sub-first}
\end{subfigure}


\begin{subfigure}{.5\textwidth}
  \centering
  \includegraphics[width=1.0\linewidth]{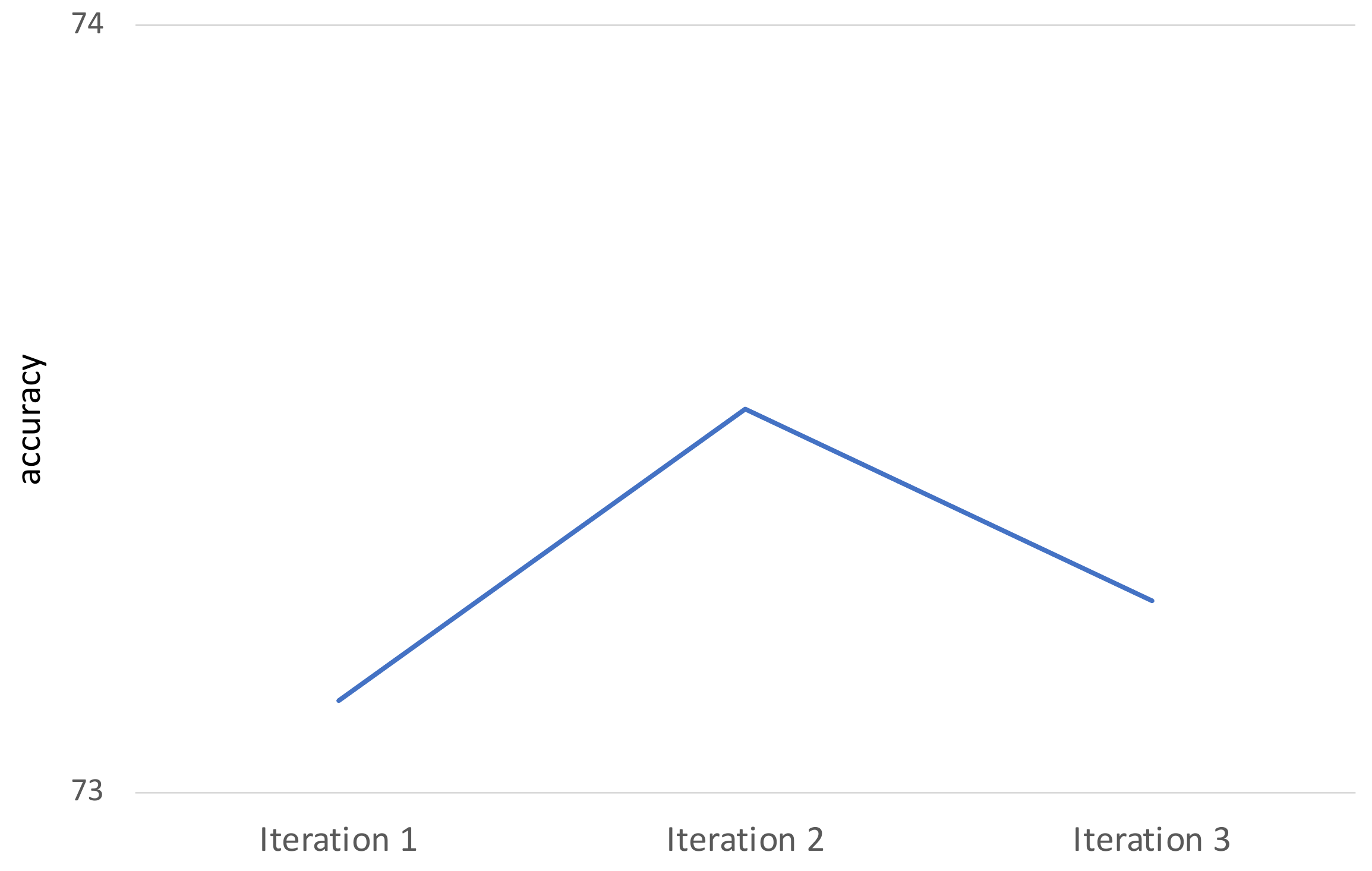}  
  \caption{ResNet-34 on FER2013. }
  \label{fig:sub-third}
\end{subfigure}
\caption{Performance improvement of PASM with changing the iteration number.  ~\LL{Based on the observations, we can find that: 1) on RAF-DB dataset, the accuracy increases with the changing iteration numbers, but the rate of increase is decreasing; 2) on FER2013, the accuracy increases first, and after two iterations of refinement, the performance starts to drop (but still higher than the first round).}}
\label{fig:page_11}
\end{figure}

\textbf{Analysis of Self Mining Iterations}
In Fig.~\ref{fig:page_11}, we report the impact of self-mining iteration numbers in our proposed method. As illustrated in Section.~\ref{sec:method}, PASM can be conducted in an iterative way. {The student network in the current iteration changes its role to a teacher network in the following iteration, generating new studying materials and providing guidance to the new student network. We report the performances of a ResNet-34 on RAF-DB and FER2013 in different iterations}. Based on the experimental results, we can find that: 1) on RAF-DB dataset, the accuracy rates increase with the changing iteration numbers, but the rate of increase decreases; 2) on FER2013, the accuracy rates increase first; after two iteration refinements, the performance starts to drop (but still higher than the first round).


\begin{figure}[htbp]
  \centering
  \includegraphics[width=1.0\linewidth]{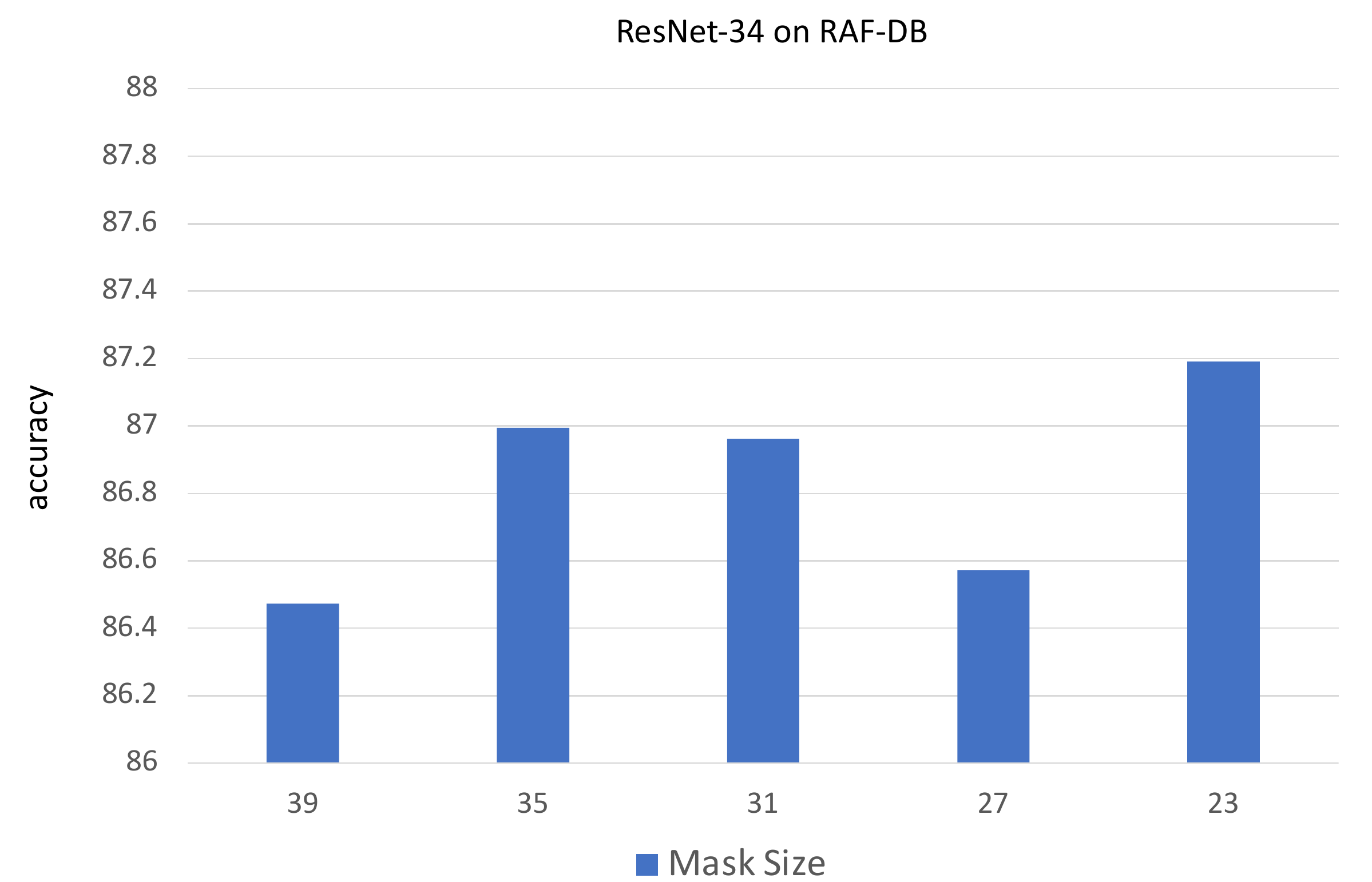}  
  \caption{ Evaluation  with  different mask sizes. ResNet-34 trained on RAF-DB. }
  \label{fig:mask_size}
\end{figure}

\textbf{Analysis of Mask Sizes and Parameter Configuration in Equation.~\ref{eq:teacher_student}}
In Fig.~\ref{fig:mask_size}, we report the prediction accuracy with different mask sizes, which is trained on RAF-DB by a ResNet-34 for one round. We choose five different mask sizes,~\textit{i.e.}, from 23 to 39 with a step size of 4. In the figure, we can find that the accuracy is not very sensitive to the mask sizes.


\begin{figure}[htbp]
  \centering
  \includegraphics[width=1.0\linewidth]{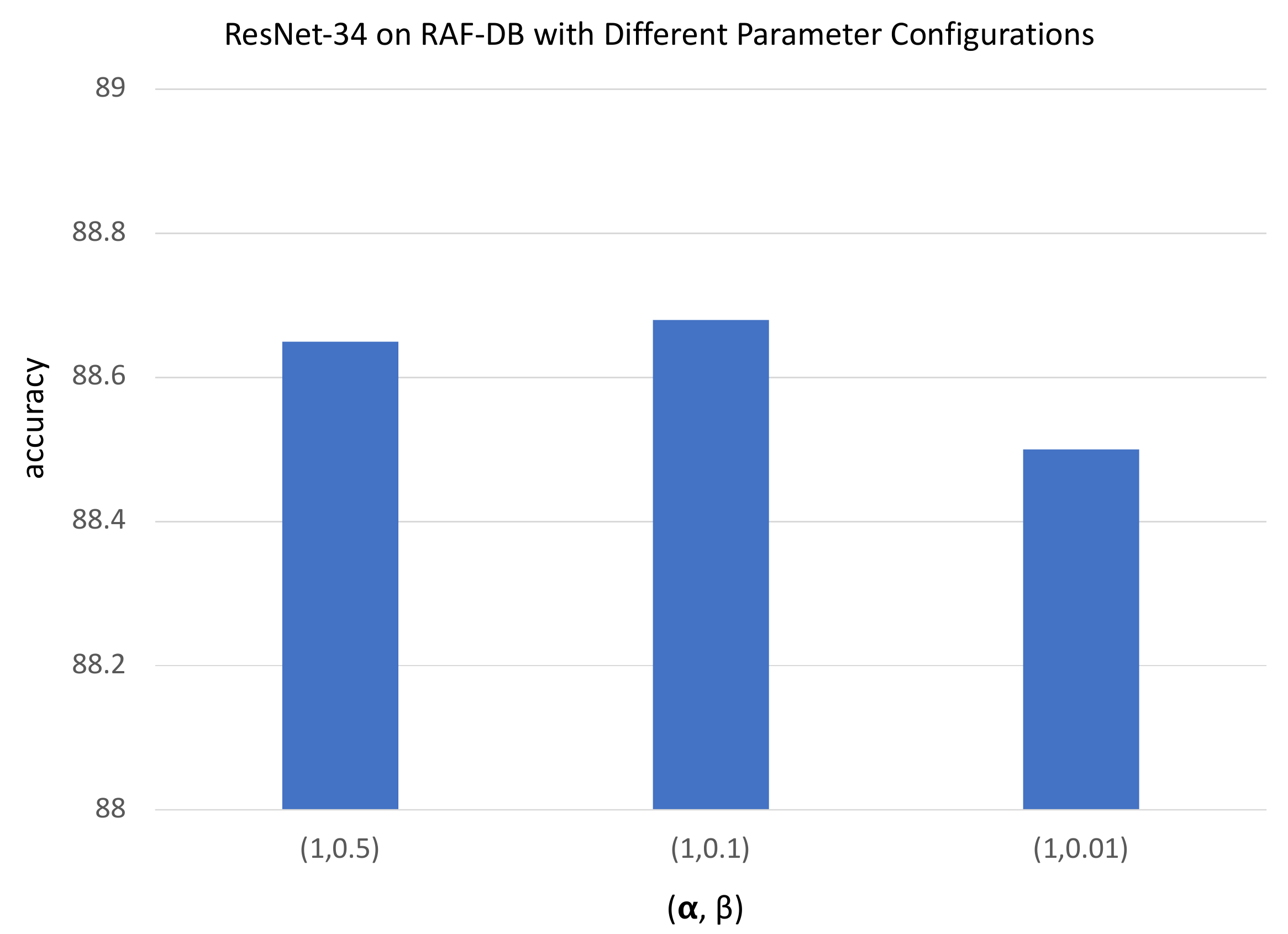}  
  \caption{ Evaluation  with  different parameter configurations in Equation.~\ref{eq:teacher_student}. ResNet-34 trained on RAF-DB. }
  \label{fig:parameter_config}
\end{figure}

In Fig.~\ref{fig:parameter_config}, we show the prediction accuracy with parameter configuration in Equation.~\ref{eq:teacher_student}. In this experiment, we train on RAF-DB via a ResNet-34 for one round. We set $\alpha$ as 1, change $\beta$ to 0.5, 0.1, 0.01 respectively. In the figure, we can find that the accuracy is stable with different parameter configurations.


\textbf{Evaluation of Random Erasing Values}
{We conduct another experiment about using random values for erasing. Specifically, we generate random values between $0$ and $255$ and use the generated random values to replace the original pixels in the erasing region. The experiment is conducted on RAF-DB with a ResNet-34, trained for one round, without teacher guidance but with new learning samples. The accuracy is $86.93\%$, which is higher than Random Erasing ($85.93\%$), comparable with PASM without teacher reported in Table.~\ref{table:random-erasing}($86.86\%$). This matches our expectation: 1) generating new learning materials is important to improve network capability; 2) when generating new learning materials to benefit PASM learning processes, how to locate the position adaptively is more important than how to process the located position.
}

\textbf{Open Problems in PASM}
There are two open problems that need to solve in the future: 1) ~\LL{the sensitive position search, as an adversarial attack in essential, might bring additional computation cost in training (only), while in inference stages, our method does not bring any additional parameters and computation cost. Currently, the point adversarial attach step consumes around 10 seconds for a 256x256 image.} This computation cost issue in adversarial learning has already aroused research attention in recent works such as~\cite{wong2020fast}, and it should be mitigated by a more advanced adversarial attack solution in the future.  2) an automatic and adaptive way for choosing an appropriate iteration number in PASM is expected. We leave those to our future work.

\begin{table}[htbp]

\caption{Study of Self Mined Information Generality}


\label{table:study-of-generality}
\centering
\begin{tabular}{lccc}
\toprule
Dataset      & Mining Network           &Training Network      & Accuracy       \\ \midrule 
PAT\cite{cai2018probabilistic} &  -   & ResNet-34  & $83.83\%$ \\
PAT\cite{cai2018probabilistic}  &  -  & VGG-16 &$84.19\%$ \\
OADN\cite{ding2020occlusion}  &  -  & ResNet-50 &$87.16\%$ \\
PASM   & ResNet-34    & ResNet-34  & $\textbf{87.54}\%$             \\ 
PASM   & ResNet-34    & VGG-16  & $\textbf{87.09}\%$             \\ \midrule
PAT\cite{cai2018probabilistic} & -    & ResNet-34  &$72.16\%$ \\
PAT\cite{cai2018probabilistic}  & -   & VGG-16 &$72.00\%$ \\
PASM   & ResNet-34    & ResNet-34  & $\textbf{73.50}\%$            \\  
PASM   & ResNet-34    & VGG-16  & $\textbf{72.73}\%$             \\  


\bottomrule
\end{tabular}

\end{table}


\section{Conclusion}
In this article, we propose a self mining framework named point adversarial self mining to improve the performance of facial expression recognition. By progressively generating new learning materials and providing guidance from an updated teacher network, the recognition capability of the student network with the same architecture can be improved in an iterative manner. Experimental results on benchmark facial expression datasets have demonstrated the efficacy of the proposed method. ~\LL{In the future, we plan to explore the possibility to apply our method on other facial activity analysis problems, such as facial action unit analysis.}

\section{Acknowledgements}
This work was supported by the Agency for Science,
Technology and Research (A*STAR) under AI and ANALYTICS SEED GRANT (Grant No. Z20F3RE005) and AME Programmatic Funding Scheme (Project No. A18A1b0045).


%





\ifCLASSOPTIONcaptionsoff
  \newpage
\fi

\bibliographystyle{IEEEtran}
\bibliography{bare_jrnl}

\end{document}